\documentclass[journal,twocolumn]{IEEEtran}
\IEEEoverridecommandlockouts                              

\usepackage{myStyle}

\begin{document}

\title{
Interaction-Aware Sampling-Based Motion Planning for Autonomous Surface Vessels in Urban Waterways
}

\author{Elia Trevisan$^{1,\dagger}$, Alejandro Gonzalez-Garcia$^{2,\dagger}$, Jan Swevers$^2$ and Javier Alonso-Mora$^1$ 
\thanks{This research is supported by the project ``Sustainable Transportation and Logistics over Water: Electrification, Automation and Optimization (TRiLOGy)'' of the Netherlands Organization for Scientific Research (NWO), domain Science (ENW), and the Amsterdam Institute for Advanced Metropolitan Solutions (AMS) in the Netherlands. A. Gonzalez-Garcia is supported by the Flanders Make SBO project ARENA (Agile \& Reliable Navigation).}
\thanks{$^{1}$Cognitive Robotics Department,
        TU Delft, 
        {\tt\small \{e.trevisan, j.alonsomora\}@tudelft.nl}}
\thanks{$^{2}$MECO Research Team, Department of Mechanical Engineering, KU Leuven, Belgium and Flanders Make @ KU Leuven, Belgium, {\tt\small \{alex.gonzalezgarcia, jan.swevers\}@kuleuven.be}}
\thanks{$^\dagger$E. Trevisan and A. Gonzalez-Garcia are equal contributors to this work.}}

\maketitle
\thispagestyle{empty}

\begin{abstract}
Autonomous surface vessels operating in urban waterways must coordinate with other vessels in confined spaces while complying with traffic regulations, often without knowing their intentions. We present a decentralized motion planner based on model predictive path integral control, a sampling-based model predictive control scheme. The planner jointly computes a trajectory for the ego vessel and a prediction of the trajectories of the other vessels, and therefore accounts for their interaction. Traffic regulations are encoded as discontinuous costs, which the sampling-based formulation admits without a smooth approximation. We evaluate the method in maps taken from real sections of Amsterdam's canals, in encounters between two vessels and in scenarios with four interacting vessels. Across 780 randomized runs the method reaches every goal without collisions or deadlocks, and outperforms an optimization-based planner designed for the same domain, as well as a version of the proposed planner that does not account for the interaction. We further validate the planner on two scaled vessel prototypes in an outdoor environment, where planning and localization run onboard at 10 Hz, and head-on and crossing encounters are resolved without collisions or rule violations. Code and videos: \href{https://autonomousrobots.nl/paper_websites/ia-mppi}{https://autonomousrobots.nl/paper\_websites/ia-mppi}.
\end{abstract}



\section{Introduction}

As urban populations continue to grow, cities are compelled to improve their transportation and mobility infrastructure. In the case of the European continent, inland waterways provide an underutilized option for urban mobility and freight transport solutions. For instance, the City of Amsterdam has the opportunity to leverage its urban canals to transport goods, waste, and people, decreasing congestion and dependence on road infrastructure~\cite{gemeente_amsterdam_uitvoeringsplan_2020}. Autonomous Surface Vessels (ASVs) can be deployed on such urban waterways to enhance the safety and efficiency of operations, as was the objective of the Roboat project~\cite{leoni_roboat_2022}. Furthermore, inland waterways also offer an underutilized, sustainable option for transporting goods and people across Europe, where ASVs could also help increase safety and reduce operational errors \cite{Domenighini2024,ZHANG2025121592}. 
Nevertheless, the nature of inland waterways poses a complex challenge to the development and deployment of ASVs. The confined spaces within narrow canals, complex vessel dynamics, floating and static obstacles, as well as other vessels, create a complex, unstructured environment. In the case of other waterway users, the ASVs must be able to safely interact with both human-piloted and autonomous agents, while obeying traffic regulations~\cite{koninkrijksrelaties_binnenvaartpolitiereglement_2017}. 

Sampling-based Model Predictive Control (MPC), and in particular Model Predictive Path Integral Control (MPPI) \cite{Williams2015}, solves motion planning problems with complex dynamics and non-smooth cost functions by rolling out and weighting a large number of sampled input sequences. In contrast to optimization-based methods, which often depend on convex approximations of free space and simplify collision avoidance by inflating the ego and obstacle agents into ellipsoids \cite{de_vries_regulations_2022,11247574}, MPPI can handle the exact shapes of both static and moving obstacles, even when these shapes are non-convex, and it admits the discontinuous costs that encode traffic rules. These properties suit the narrow canals and close encounters of inland waterways. Safe navigation among other vessels also depends on cooperation~\cite{trautman_robot_2013}. A vessel that plans around a prediction of the other vessels resolves the encounter on its own, and the maneuver it chooses may not be consistent with the maneuvers that the other vessels choose. In contrast, a vessel that accounts for how the other vessels respond to its own motion can anticipate a joint maneuver. Therefore, we present Interaction-Aware MPPI (IA-MPPI), a decentralized version of MPPI in which each vessel plans jointly for itself and for the vessels around it, and so anticipates their maneuvers without exchanging plans. The contributions of this paper are:


\begin{figure}[tb]
    \centering
    \includegraphics[width=0.45\textwidth]{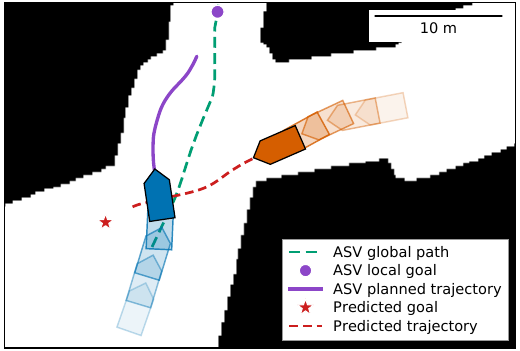}
    \caption{Crossing at an intersection in simulation. The ego vessel (blue) runs the proposed planner; the other vessel (orange) does not communicate. Faded outlines show the preceding motion. The ego vessel knows its own state, a global path (dashed green) and the position and velocity of the other vessel. It places a local goal (purple circle) ahead of itself on the global path, estimates the goal of the other vessel (red star) with a constant velocity model, and samples input sequences over the horizon for both vessels. This yields its own planned trajectory (solid purple) and the trajectory it expects from the other vessel (dashed red); only the first input of its own sequence is applied.}
\label{img:illustration}

    \label{img:illustration}
\end{figure}


 
\begin{itemize}
    \item We present a decentralized architecture for interaction-aware motion planning that requires minimal or no communication between agents.
    \item We propose a cost function design that encodes navigation objectives, allowing the system to comply with traffic regulations during operation.
    \item We validate the planner outdoors on two ASV prototypes, with planning and localization running onboard.
    \item We release an open-source implementation of the method, which is portable across platforms and is intended to support its adoption on other autonomous systems.
\end{itemize}
 
We evaluate the method in maps taken from real sections of Amsterdam's canals. In two-agent encounters we compare against a regulations-aware optimization-based MPC~\cite{de_vries_regulations_2022}, and in scenarios with four interacting vessels we compare the decentralized variants against a centralized one. Across 780 randomized runs the method reaches every goal without collisions or deadlocks. In the field, two prototypes resolved twenty head-on and crossing encounters while sustaining a 10~Hz control frequency on their onboard computers. This paper builds on our previous work~\cite{streichenberg_multi-agent_2023} in three aspects. The sampling distribution of the planner is informed by ancillary controllers, following the biased formulation of~\cite{biasedmppi}, so that distinct maneuvers are represented in the sample set. The simulation study is repeated with the resulting planner and extended with a comparison against the formulation that does not account for the interaction. The planner is further validated in field experiments with two ASV prototypes, and its implementation is released as open-source software.

The remainder of the paper is organized as follows. Section~\ref{sec:prevwork} reviews related work. Section~\ref{sec:preliminaries} introduces the sampling-based control scheme and the vessel model on which the planner is built, and Section~\ref{sec:method} presents the planner, from its decentralized architecture to the sampling scheme and the cost formulation. The planner is evaluated in simulation in Section~\ref{sec:simulation} and with real world experiments in Section~\ref{sec:realworld}. Section~\ref{sec:discussion} discusses what the two evaluations show, and Section~\ref{sec:conclusions} concludes the paper.

\section{Related Work} \label{sec:prevwork}

Cooperative and interactive motion planning is a challenging problem for autonomous vehicles, which must deal with other traffic participants as well as static and dynamic obstacles~\cite{schwarting_planning_2018}. Common approaches for different types of robots, including the Dynamic Window Approach (DWA)~\cite{fox_dynamic_1997}, Reciprocal Velocity Obstacle (RVO) \cite{van_den_berg_reciprocal_2008}, Optimal Reciprocal Collision Avoidance (ORCA)~\cite{van_den_berg_reciprocal_2011}, and Artificial Potential Fields (APF)~\cite{Khatib, ji_path_2017,LANGXIONG2025119928} have provided solutions that are highly computationally efficient.  Nevertheless, these methods are designed to passively react to other agents, which may not be ideal for highly interactive urban environments on water and ground. For instance, \cite{LANGXIONG2025119928} proposed an APF algorithm for vessels on narrow inland waterways. Simulations demonstrated various scenarios in which the vessel successfully avoided collisions with other vessels. Similar adaptations to maritime traffic rules fuse velocity obstacles with DWA~\cite{CHEN2026126631}, pair a graph search with a local reactive planner~\cite{10013088}, or account for encounter uncertainty through probabilistic velocity obstacles~\cite{YUAN2026126140}.

Learning-based approaches have also been proposed for autonomous agents on dense environments. Model-free Reinforcement Learning (RL) algorithms leverage neural network architectures and simulation environments to train policies based on reward functions. For instance, RL agents have been successfully trained to navigate ground robots interacting with crowds \cite{chen_decentralized_2017,chen_socially_2017}. However, common RL approaches cannot generalize to different systems and do not provide theoretical collision avoidance guarantees. Planning and learning were integrated in~\cite{Cai2023} for autonomous driving, where learned policy and value networks act as heuristics to a motion planner, improving computational performance at the cost of sensitivity to model error. In the maritime domain, reinforcement learning has been combined with model-based control to obtain COLREGs-compliant avoidance~\cite{10587203}, and transfer learning has been used to reduce the retraining required when the scenario changes~\cite{11106363}. Imitation Learning (IL) has also been proposed for socially compliant robot navigation. In~\cite{Kretzschmar}, it was used to model human behavior. However, an extensive optimization-based solver is still used to compute a feasible trajectory. Inverse RL instead assumes the expert optimizes a cost function learned from demonstrations, which admits a broader set of behaviors than IL. The DriveIRL planner~\cite{Phan-Minh2023} learned from a large driving dataset and was validated in real traffic. A related formulation was introduced for urban vessels in~\cite{park_social_2021}, learning behaviors from a dataset of vessel trajectories. Simulations in Amsterdam canal segments showed fewer situations requiring avoidance maneuvers, though the approach was not validated on the water.

Game-theoretic formulations model the encounter as a dynamic game between the agents, in which each one optimizes its own objective while accounting for the decisions of the others. Solvers based on iterative linear-quadratic approximations make such games tractable in real time for nonlinear dynamics and general-sum objectives~\cite{fridovich-keil_efficient_2020}. Since the objectives of the other agents are rarely known, they can be inferred from observed behavior~\cite{peters2023online}, or treated as uncertain. For instance, contingency games plan a set of trajectories conditioned on several hypotheses about the intent of the other agent, and commit to a common initial plan until that uncertainty is expected to resolve~\cite{10400882}. Bounded rationality has also been modeled directly, using level-$k$ reasoning to represent driver behavior at unsignalized intersections~\cite{tian_game-theoretic_2022}. These formulations require a parametrization of the objectives of the other agents, and their cost grows quickly with the number of agents and of hypotheses.

Model Predictive Control is a widely used approach for local motion planning in mobile robots. MPC optimizes control actions by using the robot's dynamic model to predict the system's future behavior over a finite time horizon. MPC planners can handle complex system dynamics and actuator constraints, as well as collision avoidance constraints. However, MPC planners are commonly computationally expensive, which scales in multi-agent settings \cite{9904960}. Additionally, common multi-agent settings require the ego-agent to know the actions of the others to plan around them, which is not ideal in scenarios where communication is infeasible. For instance, in \cite{luis_online_2020}, a distributed MPC was developed for motion planning with multiple drones with ideal communication conditions. Alternatively, other approaches have opted to implement constant velocity models to predict future motion, relying on an estimation of the current state of the neighbors ~\cite{de_vries_regulations_2022}, or by predicting neighbor trajectories with learning-based techniques~\cite{zhu_learning_2021}. For vessels, the intent of the encountered ship has also been inferred explicitly, through Bayesian inference over its planned route~\cite{MAHIPALA2025121564} or by including its maneuvering behavior in the risk assessment~\cite{11181576}. However, as long as planning is done based on predictions, large parts of the state-space can be perceived as unsafe by the motion planner~\cite{trautman_unfreezing_2010}. 

Traffic rules have also been encoded as hard constraints rather than costs. In~\cite{10433938}, the relevant COLREGs are formulated as affine constraints within a model predictive contouring controller, which yields a convex search space and strict compliance at the cost of a more conservative set of admissible maneuvers. Related formulations address compliance in restricted waters~\cite{HE2025121966} and the risk-dependent departures from the rules permitted by Rule 2b~\cite{11017509}.  In contrast, modeling the interaction between agents could overcome such an issue, allowing the ego-agent to be aware of how its actions influence the actions of its neighbors. Unfortunately, constrained optimization approaches can become significantly computationally expensive if they incorporate interaction models~\cite{sadigh_planning_2016}. Our formulation instead penalizes violations, which lets the planner trade rule compliance against progress when no compliant maneuver exists.

MPPI is a sampling-based method that rolls out a large number of input sequences through the system model and computes the input as a cost-weighted average of the samples. It provides approximate solutions to nonlinear, non-convex and discontinuous Stochastic Optimal Control (SOC) problems~\cite{kappen_path_2005,Williams2015,williams_information-theoretic_2018}, and in contrast to other SOC approaches such as iterative Linear Quadratic Gaussian (iLQG)~\cite{li_iterative_2007} or Differential Dynamic Programming (DDP)~\cite{Theodorou2010a}, it requires neither linearization of the dynamics nor a quadratic approximation of the cost. This makes it suitable for interaction-rich environments, where the cost encodes traffic rules and the exact shapes of the obstacles. Since the rollouts are independent, they can be evaluated in parallel, which allows the method to run in real time. Two aspects, however, limit its use in the multi-agent setting we consider. First, the samples are commonly drawn from a Gaussian centered on the previously computed input sequence, which restricts exploration to its neighborhood and slows the transition between distinct maneuvers~\cite{carius_constrained_2021,biasedmppi}. Second, when applied to centralized multi-agent architectures, the number of samples required grows notably with the number of agents~\cite{williams_model_2017}.

\section{Preliminaries}\label{sec:preliminaries}

This section introduces the sampling-based control scheme on which the proposed method is built. We first summarize the classical MPPI algorithm, then describe the biased formulation that allows samples to be drawn from an arbitrary distribution. Since MPPI requires a model of the system to perform the rollouts, we then define the dynamics of the vessels and the way the multi-agent state is assembled.

\subsection{MPPI Algorithm}\label{sec:mppi_basic}
We adopt the information-theoretic formulation of MPPI~\cite{williams_information-theoretic_2018}, which approximates a solution to stochastic optimal control problems for discrete-time systems of the form
\begin{equation}
    \mathbf{q}_{t+1} = \mathcal{F}(\mathbf{q}_t, \tilde{\mathbf{u}}_t),   \quad   \tilde{\mathbf{u}}_t \sim \mathcal{N}(\mathbf{u}_t,\mathbf{\Sigma}),
\end{equation}
where $\mathbf{q}$ is the state, $t$ the time step, $\mathcal{F}$ a nonlinear state transition function, $\mathbf{u}$ the commanded input and $\tilde{\mathbf{u}}$ the applied input, corrupted by noise of variance $\mathbf{\Sigma}$. At every time step the algorithm draws $K$ input sequences $\tilde{U}_k, \, k\in [1,K]$ from $\mathcal{N}(\mathbf{u}_t, \nu \mathbf{\Sigma})$, where $\nu$ scales the exploration~\cite{williams_model_2019}, and propagates each of them through $\mathcal{F}$ over a horizon $T$ to obtain the state trajectories
\begin{equation}
Q_k = \big[\mathbf{q}_0, \, \mathcal{F}(\mathbf{q}_0, \tilde{\mathbf{u}}_{k,0}), \, \dots, \, \mathcal{F}(\mathbf{q}_{k, T-1}, \,\tilde{\mathbf{u}}_{k,T-1})\big].
\end{equation}
Every trajectory is assigned a total cost $S_k$, composed of a stage cost and a terminal cost. The samples are then combined through \textit{importance sampling} weights, obtained from the cost of sample $k$ relative to the lowest cost among the samples $S_{min}$,
\begin{equation}\label{eq::weights}
    w_k = \frac{1}{\eta} \exp\biggl( \frac{-1}{\lambda}(S_k - S_{min})\biggr), \quad \sum_{k = 0}^{K-1} w_k  = 1,
\end{equation}
with normalization factor $\eta$ and inverse temperature $\lambda$. The weighted average of the sampled sequences,
\begin{equation}
    U^* =  \sum_{k = 0}^{K-1} w_k \tilde{U}_k,
\end{equation}
approximates the optimal input sequence, and its first element $u_0^*$ is applied to the system. The procedure is repeated at the next time step, warm-started with a time-shifted copy of $U^*$.

\subsection{Biased-MPPI}\label{sec:biased_mppi}
The derivation above requires the samples to be drawn from a Gaussian centered on the previously computed input sequence. This restriction limits the algorithm to exploring the neighborhood of the previous solution, which slows down the reaction to unexpected changes in the environment and makes it difficult to escape local minima~\cite{biasedmppi}. Both effects are relevant in our setting, where a vessel may need to abandon a plan of crossing first in favor of yielding as soon as the intention of another vessel becomes apparent.

Biased-MPPI~\cite{biasedmppi} removes this restriction. Let $p$ denote the density of the uncontrolled system and $q_s$ the density of an arbitrary sampling distribution $\mathbb{Q}_s$. Redefining the cost of a sampled input sequence as
\begin{equation}\label{eq::biased_cost}
    \tilde{S}_k = S_k + \lambda \log \biggl( \frac{p(\tilde{U}_k)}{q_s(\tilde{U}_k)} \biggr)
\end{equation}
and repeating the derivation of Section~\ref{sec:mppi_basic} causes the ratio between the two densities to cancel, so that the weights retain the form of~\eqref{eq::weights} while the samples may be drawn from $\mathbb{Q}_s$. The free energy is then bounded by the expected cost plus $\lambda$ times the Kullback-Leibler divergence between the controlled distribution and $\mathbb{Q}_s$, which means that the resulting input sequence is biased towards the sampling distribution. The magnitude of this bias decreases as the number of samples grows. We refer the reader to~\cite{biasedmppi} for the complete derivation. The practical consequence is that any set of input sequences can be inserted among the samples, provided that the associated bias is acceptable. 

The inverse temperature $\lambda$ determines how sharply the weights concentrate on the lowest-cost samples, and an appropriate value depends on the magnitude of the costs, which varies with the scenario. Following~\cite{biasedmppi}, we adapt it online from the normalization factor $\eta$, which indicates how many samples carry a significant weight,
\begin{equation}\label{eq::lambda_tuning}
\lambda_{t+1} =
\begin{cases}
    0.9\,\lambda_t & \text{if } \eta > \eta_{\text{max}}\\
    1.2\,\lambda_t & \text{if } \eta < \eta_{\text{min}}\\
    \lambda_t & \text{otherwise,}
\end{cases}
\end{equation}
which keeps the number of contributing samples between $\eta_{\text{min}}$ and $\eta_{\text{max}}$.

\subsection{Vessel State and Dynamics}\label{sec:state_dyn}

The dynamics of an ASV are described by a nonlinear second-order system~\cite{wang_design_2018,wang_roboat_2019,wang_roboat_2020}:
\begin{equation}
\begin{aligned}
    \dot{\mathbf{x}} &= \mathbf{R}\left(\phi\right)
    \mathbf{v},\\
    \mathbf{\tau} &= \mathbf{M}\dot{\mathbf{v}} + \mathbf{C}\left(\mathbf{v}\right) \mathbf{v} + \mathbf{D}\left(\mathbf{v}\right) \mathbf{v},
\end{aligned}
\end{equation}
where $\mathbf{x} = \begin{bmatrix} x &  y & \phi \end{bmatrix}^\top$ collects the position and orientation, $\mathbf{v} = \begin{bmatrix} u &  v & r \end{bmatrix}^\top$ the corresponding velocities, and $\mathbf{R}\left(\phi\right)$ maps the body frame to the inertial frame:
\begin{equation}
    \mathbf{R}\left(\phi\right) = \begin{bmatrix}
        \cos \phi & -\sin \phi & 0 \\
        \sin \phi & \cos \phi & 0 \\
        0 & 0 & 1
    \end{bmatrix}.
\end{equation}
The matrix $\mathbf{M} \in \mathbb{R}^{3\times3}$ collects the vessel mass and the added mass,
\begin{equation}
    \mathbf{M} = \text{diag}\{m_{11},m_{22},m_{33}\},
\end{equation}
$\mathbf{C}\left(\mathbf{v}\right) \in \mathbb{R}^{3\times3}$ collects the rigid-body and added-mass Coriolis terms, written here for an origin that coincides with the center of mass of the vessel,
\begin{equation}
    \mathbf{C}\left(\mathbf{v}\right) = \begin{bmatrix}
        0 & 0 & -m_{22}v \\
        0 & 0 & m_{11}u \\
        m_{22}v & -m_{11}u & 0
    \end{bmatrix},
\end{equation}
and $\mathbf{D}\left(\mathbf{v}\right) \in \mathbb{R}^{3\times3}$ accounts for linear damping,
\begin{equation}
\mathbf{D}\left(\mathbf{v}\right) = \text{diag}\{X_{u},Y_{v},N_{r}\}.
\end{equation}
We consider the Quarterscale Roboat ASV~\cite{wang_roboat_2019}, actuated by four thrusters $(f_1,f_2,f_3,f_4)$, so that the vector of forces and moments $\mathbf{\tau} \in \mathbb{R}^{3\times1}$ follows from the thrust allocation
\begin{equation}
    \mathbf{\tau} = \mathbf{B}\mathbf{u} = \begin{bmatrix}
        1 & 1 & 0 & 0 \\
        0 & 0 & 1 & 1 \\
        \dfrac{a}{2}&-\dfrac{a}{2} & \dfrac{b}{2} &-\dfrac{b}{2}
    \end{bmatrix}
    \begin{bmatrix}
        f_1\\
        f_2\\
        f_3\\
        f_4
    \end{bmatrix},
\end{equation}
with $a$ the distance between the port and starboard propellers and $b$ the distance between the fore and aft propellers.

Following the notation of~\cite{williams_model_2017}, the state of agent $i$ concatenates its position, heading and associated velocities,
\begin{equation}\label{eq::dynamics}
    \mathbf{q}_i = \begin{bmatrix}  \mathbf{x}_i^\top &
                                    \mathbf{v}_i^\top \end{bmatrix}^\top,
\end{equation}
and the state of the full system is obtained by stacking the states of every agent in the set $\mathcal{M}=\{0,1,...,m\}$,
\begin{equation} \label{eq::state}
    \mathbf{q} = \begin{bmatrix}
    \mathbf{q}_0^\top &
    \mathbf{q}_1^\top &
    \hdots &
    \mathbf{q}_m^\top
    \end{bmatrix}^\top.
\end{equation}

\subsection{Global Planning} \label{sec:globalplan}
On large maps the ego-agent is provided with a global path, computed with the path planning plugin of the ROS navigation stack~\cite{marder-eppstein_office_2010}. Rather than tracking this path directly, we select a local goal $\mathbf{p}_g$ on it at a look-ahead distance $r_{\textbf{p}_g}$ (Fig.~\ref{img:local_goals}). Tracking a local goal instead of the path itself leaves the local planner free to deviate when a collision avoidance maneuver requires it.

\section{Decentralized Interaction-aware Model Predictive Path Integral Control} \label{sec:method}
This section presents the proposed method. We first describe the decentralized architecture, then the two mechanisms that shape the sample set, namely the two-stage evaluation and the sampling from ancillary controllers. We then define the cost function that encodes the navigation task and the navigation rules, and finally the prediction of the local goals of the other vessels.

\begin{figure}[t]
    \centering
    \medskip
    \medskip
    \includegraphics[width=0.4\textwidth]{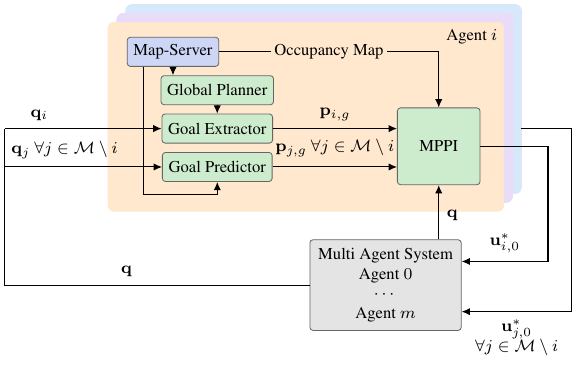}
    \caption{Structure of the planner. At every time step each agent observes the states of all agents in the scene. The local goal of the ego agent (agent $i$ in the figure) is taken from the global planner, while the local goals of the remaining agents are estimated from their states and the occupancy map. The resulting multi-agent problem is solved as if it were centralized, producing an input sequence and a trajectory for every agent, of which only the first input $\mathbf{u}^*_{i,0}$ is applied to the ego agent. The panels behind represent the other agents, each running the same structure on its own state.}
    \label{img:overview}
\end{figure}

\subsection{Approach and Architecture}\label{sec:architecture}
Each agent runs its own MPPI solver over a local multi-agent system, which allows it to anticipate the actions of the other agents (see Fig.~\ref{img:overview}). For agent $i$, the state and control output of the solver are
\begin{equation}
\begin{aligned}
    \mathbf{q}^i&=\begin{bmatrix}\mathbf{q}_i^\top& \mathbf{q}^{i\top}_j\end{bmatrix}^\top,\\
    \mathbf{u}^i&=\begin{bmatrix}\mathbf{u}_i^\top& \mathbf{u}^{i\top}_j\end{bmatrix}^\top,
\end{aligned}
\quad\forall j\in\mathcal{M}\setminus i,
\end{equation}
where $\left(.\right)^i_j$ denotes a quantity that agent $i$ estimates of agent $j$. In the centralized case of Section~\ref{sec:mppi_basic} the system state is fully observable and every agent executes the input computed by the central solver. In the decentralized case the positions and velocities of the other agents have to be observed or communicated, and although each agent samples control actions for all agents, only its own action is executed. There is therefore no guarantee that the other agents behave as predicted.

\begin{figure}[t]
    \centering
    \medskip
    \includegraphics[width=0.45\textwidth]{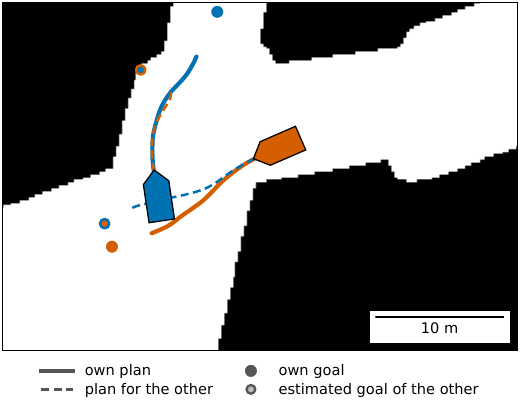}
    \caption{Two vessels  running the decentralized planner without communication. Each vessel plans its own trajectory (solid line in its color) and the one it expects from the other (dashed line in its color). Each vessel uses its own local goal (filled circle) and its estimated goal of the other vessel (outlined circle). Although the goals are estimated with a constant velocity model, both agents plan a cooperative maneuver. }    \label{img:optimization_example}
\end{figure}

Three assumptions delimit the coordination problem we address. First, the positions and velocities of the other agents are observed exactly, so that $\mathbf{q}^i_j=\mathbf{q}_j$. Second, all agents act rationally and minimize the same global cost. Third, the agents considered in our experiments are homogeneous and share the same dynamics. The method itself does not require this, since different models can be used for different agents as long as those models are known. Fig.~\ref{img:optimization_example} shows a simulated encounter between two ASVs running the proposed algorithm without communication.

\subsection{Two-stage Sample Evaluation}
As the number of agents grows, the probability that at least one of them collides with a static obstacle in a given rollout increases. In the classical implementation this causes most rollouts to receive a high cost and to be effectively rejected, so that few samples contribute to the solution. For illustration, if a fraction $p$ of the samples of each agent leads to a collision with the environment, a joint sample of all agents is collision-free with probability $(1-p)^{|\mathcal{M}|}$. With four agents and $p = 0.5$, one joint sample in sixteen survives, reducing the effective number of samples from $K$ to about $K/16$. We therefore split the sampling into two stages, as introduced in \cite{streichenberg_multi-agent_2023} and shown in Algorithm~\ref{alg:mppi_adapted_short}. In the first stage, the control samples $U_{j,k}$ of each agent $j\in\mathcal{M}$ are evaluated in parallel, producing the individual trajectories $Q_{j,k}$ and the agent-centric costs $S_{j,k}$ of~\eqref{eq::agent_cost}. Any sample whose cost exceeds the collision penalty $C_{\text{collision}}$ is discarded at this point (lines 12-13 of Algorithm~\ref{alg:mppi_adapted_short}). In the second stage, the $K$ system samples are assembled by drawing uniformly from the surviving samples of each agent and stacking them as in~\eqref{eq::state}. The complete cost $S_k$ of a system sample $Q_k$ is then obtained by summing the stored agent-centric costs $S_{j,k}$ and the costs that arise from collisions between vessels and from regulation violations (line 16, Algorithm~\ref{alg:mppi_adapted_short}).

\subsection{Sampling from Ancillary Controllers}\label{sec:ancillary}
The biased formulation of \cite{biasedmppi} allows part of the sample set to be replaced by input sequences that are not perturbations of the previous plan. We exploit this by designing a small set of ancillary controllers, each of which proposes one input sequence over the horizon at every time step. Two of them are open-loop maneuvers and two track a velocity reference:

\begin{itemize}
    \item \emph{Braking}, which commands a zero velocity reference.
    \item \emph{Go-slow}, which commands a small forward thrust.
    \item \emph{Go-fast}, which commands a large forward thrust.
    \item \emph{Go-to-goal}, which produces a velocity reference directed at the local goal at every step of the horizon.
\end{itemize}

The velocity references of the braking and go-to-goal maneuvers are converted into thrusts by a linear $\mathcal{H}_\infty$ controller, which is robust to the nonlinearities of the vessel model~\cite{biasedmppi}. These four maneuvers cover the decisions that resolve an encounter. In a crossing, a vessel either proceeds or yields, and the corresponding input sequences differ substantially from each other. Sampling only around the previous plan makes the transition between the two slow, because the samples that would realize the alternative decision are unlikely to be drawn. Including the braking and go-slow sequences ensures that a yielding maneuver is always evaluated, and including go-to-goal ensures that a direct route to the goal is always evaluated, regardless of the previous plan. The go-to-goal sequence also shortens the distance travelled, since it biases the solution towards the goal rather than towards the neighborhood of the previous solution.

Let $J$ denote the number of ancillary controllers. At every time step, the first $J$ samples of each agent are the sequences proposed by the ancillary controllers and the remaining $K-J$ samples are drawn as in Section~\ref{sec:mppi_basic}. Since $J \ll K$, the associated bias is small, and it is further reduced as $K$ grows. The bounds $\eta_{\text{min}}$ and $\eta_{\text{max}}$ used in~\eqref{eq::lambda_tuning} are reported in Table~\ref{tab::params}.

\begin{remark}
The set of ancillary controllers is not restricted to hand-designed maneuvers. Any policy that returns an input sequence over the horizon can be included, for instance a trajectory tracked from a learned prediction model~\cite{ia-mppi-lbm}, at the cost of the additional bias it introduces.
\end{remark}

\begin{algorithm}[th]

\caption{Decentralized IA-MPPI for agent $i$ (agent-specific superscripts are dropped for clarity)} \label{alg:mppi_adapted_short}

\begin{algorithmic}[1]
\Require $U$ \Comment{previous control sequence (hot start)}
\Require $\mathbf{q}$\Comment{current system state}
\State $\mathbf{p}_{i,g} \leftarrow $ \textit{receiveEgoLocalGoal}()
\State $\mathbf{p}_{j,g} \leftarrow $ \textit{predictLocalGoal}$(\mathbf{q}_j)$\Comment{$\forall \,j\in\mathcal{M}\setminus i$}
\For {each agent $j\in\mathcal{M}$}\Comment{independent rollouts}
    \For{each sample $k$}
        \If{$k \leq J$}\Comment{ancillary controllers}
            \State $\tilde{U}_{j,k} \leftarrow$ \textit{ancillaryCtrl}$_k(\mathbf{q}_j,\mathbf{p}_{j,g})$
        \Else
            \State $\mathcal{E}_{j,k} = \left[\epsilon_{0}, \, \dots, \, \epsilon_{T-1}\right]$\Comment{$\epsilon_t \in \mathcal{N}(0, \nu \mathbf{\Sigma})$}
            \State $\tilde{U}_{j,k} = U_j + \mathcal{E}_{j,k}$
        \EndIf
        \State $Q_{j,k} \leftarrow$ \textit{simulateSystem}$(\mathbf{q}_j,\tilde{U}_{j,k})$
        \State  $S_{j,k} \leftarrow$ \textit{getIndividualCost}$(Q_{j,k},\tilde{U}_{j,k},\mathbf{p}_{j,g})$
        \If{$S_{j,k} > C_{\text{collision}}$}
            \State \textit{discardSample}$(Q_{j,k},\tilde{U}_{j,k},S_{j,k})$
        \EndIf
    \EndFor
\EndFor
\State Uniformly sample from the remaining valid input sequences to rebuild $K$ full system samples
\For{$k=1:K$}
    \State $\mathcal{S}_k = \sum_{j\in\mathcal{M}}S_{j,k} +         \textit{getConfigurationCost}(Q_{k})$
\EndFor

\State $\big[w_1, \, \dots, \, w_K\big]$ = \textit{importanceSampling}$(\big[\mathcal{S}_1, \, \dots, \, \mathcal{S}_K\big])$
\State $\lambda \leftarrow$ \textit{updateInverseTemperature}$(\eta)$\Comment{\eqref{eq::lambda_tuning}}
\State return $U^{*} = \sum_{k = 1}^{K} w_k \tilde{U}_k$
\end{algorithmic}
\end{algorithm}

\subsection{Cost Formulation}
The sample cost $S_k, \forall k \in [1,K]$ is split into agent-centric and configuration costs. Both are instantaneous and are evaluated at every time step $t$ of the horizon $T$ and for every sample $k$. The \textbf{agent-centric} cost $S_{j,k,t}$, denoted $S_{\text{agent}}$ in the following, applies to agent $j$ and sample $k$ and is defined as
\begin{equation}\label{eq::agent_cost}
    S_{\text{agent}} = C_{\text{static}} + C_{\text{rotation}} + C_{\text{tracking}} + C_{\text{speed}} + C_{\text{sample}}.
\end{equation}
The term $C_{\text{static}}$ applies the constant penalty $C_{\text{collision}}$ when the vessel enters occupied space, and $C_{\text{rotation}}$ penalizes rotational velocity linearly, with scale $k_{\text{rot, slow}}$ below $\vert \vert\mathbf{v}\vert \vert_2 = 0.5 \text{m/s}$ and $k_{\text{rot}}$ above it. Progress towards the local goal is encouraged by
\begin{equation}
    C_{\text{tracking}} = k_{\text{tracking}} \frac{\vert\vert\mathbf{p}_{g} - \mathbf{p}_{t}\vert \vert_2}
    {\vert\vert\mathbf{p}_{g} - \mathbf{p}_{t_0}\vert \vert_2},
\end{equation}
with scaling factor $k_{\text{tracking}}$, local goal $\textbf{p}_g$, predicted position $\textbf{p}_{t}$ at time $t$ and position $\textbf{p}_{t_0}$ at the start of the horizon. A constant penalty $C_{\text{speed}}$ is applied when the speed exceeds the maximum, and the input is penalized by
\begin{equation}\label{eq::sample_cost}
    C_{\text{sample}}(\mathbf{u}_{t}, \mathbf{\epsilon}_{t})\ = \frac{1}{2} \gamma [ \, \mathbf{u}_{t}^T \Sigma^{-1} \mathbf{u}_{t} + 2 \mathbf{u}_{t}^T \Sigma^{-1} \mathbf{\epsilon}_{t} ],
\end{equation}
with tuning parameter $\gamma$.

\begin{figure}[t]
    \centering
    \medskip
    \includegraphics[width=0.4\textwidth]{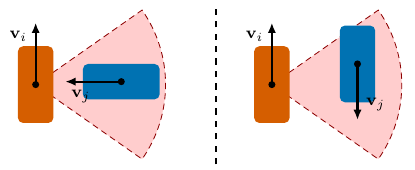}
     \caption{Encounter geometries penalized by the regulation cost. The shaded region is the starboard side of the ego vessel within the radius $k_{\text{region}}$, in which the checks apply. Left: the ego vessel does not give way to a vessel approaching from starboard. Right: the ego vessel avoids an oncoming vessel to port.}
    \label{img:regulations}
\end{figure}

The \textbf{configuration cost} $S_{k,t}$, denoted $S_{\text{configuration}}$, is evaluated at every time step $t$ of the horizon for every sample $k$ and collects the terms that depend on more than one agent,
\begin{equation}
    S_{\text{configuration}} = C_{\text{dynamic}} + C_{\text{regulation}}.
\end{equation}
Collisions between vessels are penalized by $C_{\text{dynamic}}$ with the same constant $C_{\text{collision}}$ used for static obstacles. The regulation cost $C_{\text{regulation}}$ encodes the two navigation rules that are relevant to our scenarios, namely \emph{avoiding to the right} in head-on encounters and \emph{right of way} in crossings, both derived from COLREGs~\cite{johansen_ship_2016}. Compliance is determined from the relative position and the relative velocity of a pair of vessels. For the position, we check whether a vessel with significant velocity ($\vert \vert \mathbf{v} \vert \vert > v_{\text{min}}$) lies on the starboard side within a given radius (Fig.~\ref{img:regulations}). For the velocity, a vessel is considered to approach from the right when
\begin{equation}\label{eq::row}
    \vert\vert \mathbf{v}_i \times \mathbf{v}_j \vert \vert < \vert\vert \mathbf{v}_i \vert\vert \, \vert\vert \mathbf{v}_j \vert \vert \sin(-\frac{\pi}{2} + \delta),
\end{equation}
with angular margin $\delta$, and to approach head-on when
\begin{equation}\label{eq::fs}
    \vert\vert \mathbf{v}_i \cdot \mathbf{v}_j \vert\vert < \vert\vert \mathbf{v}_i \vert\vert \cdot \vert \vert \mathbf{v}_j \vert \vert \cos(\pi + \delta).
\end{equation}
A vessel detected on the starboard side whose velocity satisfies~\eqref{eq::row} means that the ego-agent is breaking the right-of-way rule (Fig.~\ref{img:regulations} left), while a vessel on the starboard side with opposite velocity means that it is being passed on the right (Fig.~\ref{img:regulations} right).


\subsection{Local Goal Prediction} \label{sec:goalpred}
\begin{figure}[t]
    \centering
    \medskip
    \includegraphics[width=0.45\textwidth]{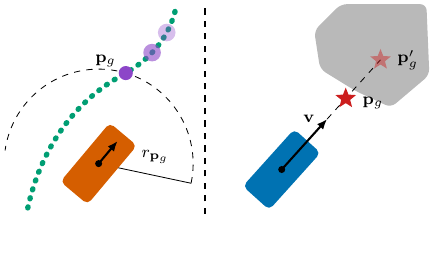}
    \caption{Left: the local goal $\mathbf{p}_g$ of the ego vessel is found by moving backward along the global path (faded circles) until the point lies within a radius $r_{\mathbf{p}_g}$ of the vessel. Right: the goal of another vessel follows from a constant velocity model; when the resulting point $\mathbf{p}_g'$ falls on a static obstacle, the goal is moved back along the velocity direction to the first free point $\mathbf{p}_g$.}
    \label{img:local_goals}
\end{figure}

The decentralized formulation requires an estimate of the local goal of every non-ego vessel (line 2, Algorithm~\ref{alg:mppi_adapted_short}). We use a constant velocity model, so that agent $i$ estimates the goal of agent $j$ as
\begin{equation}
    \mathbf{p}^i_{j,g} = T \, \delta T \, \mathbf{R}\left(\phi^i_j\right) \mathbf{v}^i_j  + \mathbf{p}^i_{j},
\end{equation}
with horizon length $T$ and step size $\delta T$, so that the vessel is projected over the duration of the horizon. When the estimated goal lies in occupied space, we move back along the vector towards $\mathbf{p}^i_j$ and take the first free point, as shown in Fig.~\ref{img:local_goals}.

\begin{remark}
A constant velocity model is a coarse estimate of intent, and it degrades when the other vessel is executing a maneuver. Learning-based trajectory predictors can be used in its place, and have been shown to reduce the goal estimation error in interaction-rich canal scenes~\cite{ia-mppi-lbm}. The proposed framework is agnostic to this choice, since the predictor only has to return a reachable position for each vessel. We retain the constant velocity model in this work because it requires no training data and no additional computation on board.
\end{remark}

\begin{remark}
The proposed interaction-aware framework is not specific to surface vessels. A portable implementation, built on a parallelized sampling-based MPC library, is available at \url{https://github.com/tud-amr/interaction_aware_mppi}. This software package includes additional examples with point and ground robots.
\end{remark}

\begin{table}[t]
\caption{Parameters used in simulation and in the real-world experiments.}
\label{tab::params}
\begin{center}
\footnotesize
\setlength{\tabcolsep}{4pt}
\begin{tabular}{c l c c}
 \toprule
 \textbf{Param.} & \textbf{Description} & \textbf{Sim.} & \textbf{Real} \\
 \midrule
 $K$ & Number of samples & 1000/2000 & 1000 \\
 $J$ & Ancillary controllers & 4 & 4 \\
 $T$ & Horizon [steps] & 100 & 100 \\
 $\delta T$ & Time step [s] & 0.1 & 0.1 \\
 $\lambda_0$ & Initial inverse temperature & 15 & 15 \\
 $\eta_{\text{min}}$, $\eta_{\text{max}}$ & Sample count bounds & 3, 6 & 3, 6 \\
 $\nu$ & Variance scaling & 12 & 6 \\
 $C_{\text{collision}}$ & Collision penalty & 2000 & 10000 \\
 $C_{\text{ROW}}$, $C_{\text{ATR}}$ & Rule penalties & 100 & 1000 \\
 $k_{\text{rot}}$ & Rotation penalty & 1.0 & 0.5 \\
 $k_{\text{rot,slow}}$ & Rotation penalty, low speed & 5.0 & 2.5 \\
 $k_{\text{tracking}}$ & Tracking cost & 3.5 & 9.0 \\
 $k_{\text{speed}}$ & Speed penalty & 3 & 3 \\
 $\gamma$ & Input cost & 0.01 & 0.01 \\
 $\delta$ & Angular margin [rad] & 1.0 & 0.52 \\
 $k_{\text{region}}$ & Rule radius [m] & 6.0 & 12.0 \\
 $v_{\text{min}}$ & Minimum rule speed [m/s] & 0.2 & 0.05 \\
 $r_{\mathbf{p}_{g}}$ & Local goal radius [m] & 15 & 5 \\
 & Maximum speed [m/s] & 1.7 & 0.3 \\
 & Maximum thrust [N] & 15 & 4 \\
 \bottomrule
\end{tabular}
\end{center}
\end{table}

\section{Simulation Experiments}\label{sec:simulation}

We evaluate the proposed method in maps taken from real sections of Amsterdam's canals, namely the Herengracht (HG), the Prinsengracht (PG), the Bloemgracht (BG) and the intersection between the Bloemgracht and the Lijnbaansgracht (BGLG). The dimensions of both the maps and the vessels are represented faithfully. Section~\ref{sec:sim_two_agent} compares our method in two-agent encounters against an optimization-based MPC and against the formulation that does not account for the interaction. Section~\ref{sec:sim_crowded} evaluates it in interaction-rich scenarios with four agents. Section~\ref{sec:sim_computation} reports how the computation time scales with the number of agents.

In every experiment we evaluate three variants of the proposed method, and a fourth variant that does not account for the interaction. In the \emph{centralized} variant a single solver plans for all vessels and each vessel executes the corresponding input. In the \emph{decentralized with communication} variant each vessel runs its own solver and the local goals of the other vessels are communicated. In the \emph{decentralized without communication} variant the local goals are instead estimated with the constant velocity model. In the \emph{decentralized with prediction} variant the other vessels enter the rollouts of each vessel as constant-velocity predictions of their observed states; the vessel samples inputs for itself only and evaluates its collision and rule costs against the predictions. The parameters are listed in Table~\ref{tab::params}. We use $K=1000$ samples in the two-agent experiments and $K=2000$ in the four-agent and scalability experiments, where the larger state space requires a denser exploration of the input space.

\subsection{Two-Vessel Encounters}\label{sec:sim_two_agent}

\begin{figure}[t]
    \centering
    \includegraphics[width=\columnwidth]{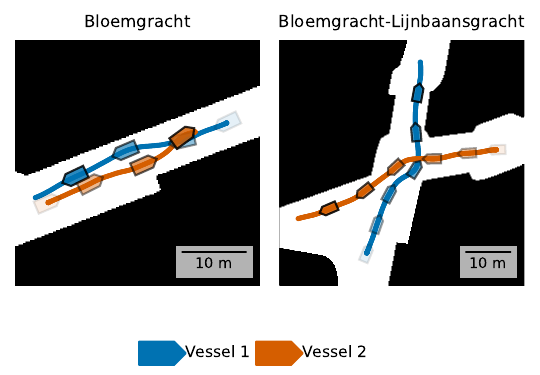}
    \caption{Crossing at the Bloemgracht-Lijnbaansgracht intersection resolved by the decentralized method without communication (left) and by the decentralized method with constant velocity model prediction (right). On the left, the give-way vessel (blue) passes behind the vessel with right of way and complies with the regulation. On the right, the vessel with right of way slows down and the give-way vessel crosses ahead of it, which does not comply with the regulation.}
    \label{img:sim_two_agent}
\end{figure}

We first consider two-agent encounters. The three interaction-aware variants, the decentralized variant with prediction and the Regulations Aware Model Predictive Contouring Controller (RA-MPCC)~\cite{de_vries_regulations_2022} are evaluated in two scenarios: a head-on encounter in BG and a crossing at the BGLG intersection. Each scenario is run 100 times with randomized initial conditions and goals, drawn from the same initialization regions for every method. Fig.~\ref{img:sim_two_agent} shows one run of each scenario.

Table~\ref{tab::two_agent} summarizes the results. The RA-MPCC results are those reported in our previous work~\cite{streichenberg_multi-agent_2023}, obtained in the same scenarios and initialization regions. We report the number of runs that ended in success, in a deadlock or in a collision. For the successful runs we report the number of rule violations, the average time until all agents reach their goal, and the total average distance, defined as the sum of the average distances traveled by the agents.

The three interaction-aware variants complete every run in both scenarios, without collisions, deadlocks or rule violations. The RA-MPCC reaches its goal in 52 of 100 runs in the head-on encounter and in 74 of 100 runs at the intersection. Its failures are dominated by deadlocks, which arise because rule compliance is encoded as a continuous cost that acts as a repulsive force between the vessels, and because the free space has to be approximated by convex regions that leave little room in the narrower sections of the canal. The proposed method instead evaluates the occupancy grid directly and penalizes rule violations with a discontinuous cost, which does not introduce such a repulsion.

The arrival times of the RA-MPCC are roughly twice those of the proposed method, while its traveled distances are shorter. Both follow from the same cause: the vessels move more slowly and take more direct paths within the convex regions available to them, at the cost of frequently failing to resolve the encounter. 

\begin{figure}[t]
    \centering
    \includegraphics[width=\columnwidth]{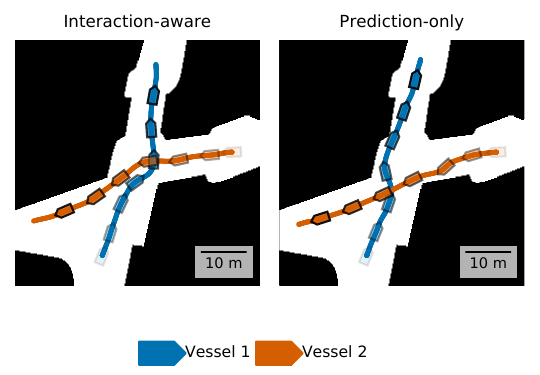}
    \caption{Crossing at the Bloemgracht-Lijnbaansgracht intersection from the same initial conditions. Left: decentralized without communication; the give-way vessel (blue) passes astern of the vessel with right of way. Right: decentralized with prediction; the vessel with right of way slows down and the give-way vessel crosses ahead of it.}
    \label{img:crossing_comparison}
\end{figure}

The decentralized variant with prediction completes 92 of the 100 head-on encounters and 98 of the 100 crossing encounters, with 48 and 11 rule violations, respectively. Each vessel expects the other to hold its course and resolves the encounter on its own. In the head-on encounters, the rules require both vessels to avoid to the right, so that they pass port to port. In a quarter of the runs, both vessels avoid to the left instead. In the crossing encounters, the rules require the give-way vessel, which has the other vessel on its starboard side, to let it pass. In most runs, the vessel with right of way yields and the give-way vessel crosses ahead of it (Fig.~\ref{img:crossing_comparison}). The interaction-aware variants evaluate the regulation cost for both vessels in the joint plan and select a maneuver that complies for both. All of these results are obtained with $K=1000$ samples. The two-vessel encounters therefore do not require a significantly dense sampling of the input space to be resolved reliably, which is relevant for deployment on embedded hardware.

\begin{table}[t]
\caption{Results of 100 runs per scenario with two agents and randomized initial conditions and goals, with $K=1000$.}
\label{tab::two_agent}
\begin{center}
\footnotesize
\setlength{\tabcolsep}{3.5pt}
\begin{tabular}{c l c c c c}
\toprule
 & \textbf{Method}
 & \shortstack{\textbf{Succ.-}\\\textbf{Dead.-}\\\textbf{Coll.}}
 & \shortstack{\textbf{Rule}\\\textbf{Viol.}}
 & \shortstack{\textbf{Avg.}\\\textbf{Time [s]}}
 & \shortstack{\textbf{Total Avg.}\\\textbf{Dist. [m]}} \\
\midrule
\multirow{5}{*}{\rotatebox{90}{BG}}
 & Centralized     & 100 - 0 - 0  & 0  & 25.44 & 76.50 \\
 & Dec. Comm.      & 100 - 0 - 0  & 0  & 25.59 & 76.83 \\
 & Dec. No Comm.   & 100 - 0 - 0  & 0  & 25.84 & 76.34 \\
 & Dec. Prediction & 92 - 2 - 6   & 48 & 29.98 & 81.34 \\
 & RA-MPCC         & 52 - 37 - 11 & 4  & 55.93 & 67.64 \\
\midrule
\multirow{5}{*}{\rotatebox{90}{BGLG}}
 & Centralized     & 100 - 0 - 0  & 0  & 28.00 & 90.37 \\
 & Dec. Comm.      & 100 - 0 - 0  & 0  & 28.00 & 90.13 \\
 & Dec. No Comm.   & 100 - 0 - 0  & 0  & 27.98 & 89.76 \\
 & Dec. Prediction & 98 - 0 - 2   & 11 & 29.83 & 83.85 \\
 & RA-MPCC         & 74 - 23 - 3  & 38 & 54.46 & 84.20 \\
\bottomrule
\end{tabular}
\end{center}
\end{table}

\subsection{Navigation in Crowded Environments}\label{sec:sim_crowded}

\begin{figure*}[t]
    \centering
    \includegraphics[width=0.9\textwidth]{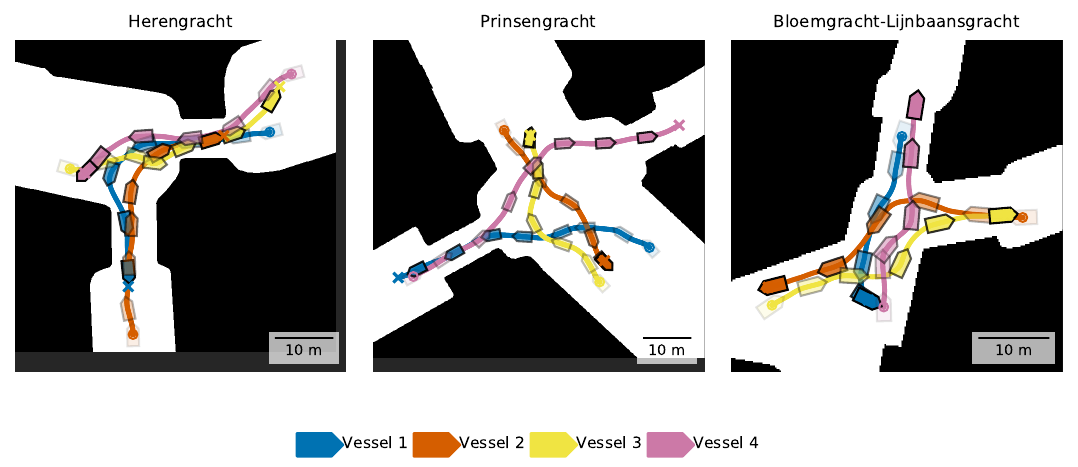}
    \caption{Four vessels navigating with the decentralized method without communication in the Herengracht, the Prinsengracht and the Bloemgracht-Lijnbaansgracht intersection. The vessel outlines are drawn at regular intervals along the trajectory and are to scale.}
    \label{img:sim_four_agent}
\end{figure*}

We next consider scenarios with four agents, which are run 20 times each with randomized initial conditions and goals in the HG, the PG and the BGLG intersection. These maps expose the vessels to repeated encounters in narrow sections and, in the case of the HG and the BGLG, under bridges. We use $K=2000$ samples. Fig.~\ref{img:sim_four_agent} shows one run in each map.

The results are summarized in Table~\ref{tab::four_agent}. The three interaction-aware variants complete all 180 runs without collisions or deadlocks, with three rule violations in total.
Decentralizing the planner has little effect when the local goals are communicated: the arrival times and traveled distances of the centralized and decentralized variants differ by less than 5\% in all three maps.

When the local goals are not communicated, the picture changes in the narrower maps. In HG, the arrival time increases from 44.70 s to 56.10 s, and the traveled distance from 233.58 m to 268.51 m. Additionally, two of the three rule violations of the whole experiment occur in this configuration. A smaller effect is visible at the BGLG intersection. In PG, which is the widest of the three maps, the three variants have similar behavior between them.

This is a consequence of the constant velocity model used to estimate the local goals of the other vessels. When four vessels perform large avoidance maneuvers in a confined space, their motion departs from a straight line, and the estimated goals diverge from the true ones. The predictions computed for the other vessels are then inaccurate, and the ego-vessel plans against a configuration that does not materialize. The effect is bounded: no run ends in a collision or a deadlock, and the cost is paid in travel time, traveled distance and occasional rule violations rather than in safety. 

The decentralized variant with a fixed prediction receives the same states of the other vessels, but does not account for the interaction. In all three maps, some of its runs end in a deadlock or a collision, 18 of the 60 in total. Its rule violations exceed one hundred, against three for the interaction-aware variants. When four vessels maneuver at once, the prediction (that each neighbor vessel holds its course) is wrong, and no vessel anticipates the maneuvers of the others. Its arrival times and traveled distances are computed over the successful runs only, and are not comparable with those of the other variants.

\begin{table}[t]
\caption{Results of 20 runs per scenario with four agents and randomized initial conditions and goals, with $K=2000$ for the MPPI-based planners.}
\label{tab::four_agent}
\begin{center}
\footnotesize
\setlength{\tabcolsep}{3.5pt}
\begin{tabular}{c l c c c c}
\toprule
 & \textbf{Method}
 & \shortstack{\textbf{Succ.-}\\\textbf{Dead.-}\\\textbf{Coll.}}
 & \shortstack{\textbf{Rule}\\\textbf{Viol.}}
 & \shortstack{\textbf{Avg.}\\\textbf{Time [s]}}
 & \shortstack{\textbf{Total Avg.}\\\textbf{Dist. [m]}} \\
\midrule
\multirow{4}{*}{\rotatebox{90}{HG}}
 & Centralized     & 20 - 0 - 0 & 0  & 44.70 & 233.58 \\
 & Dec. Comm.      & 20 - 0 - 0 & 0  & 40.50 & 226.77 \\
 & Dec. No Comm.   & 20 - 0 - 0 & 2  & 56.10 & 268.51 \\
 & Dec. Prediction & 9 - 4 - 7  & 60 & 38.50 & 211.43 \\
\midrule
\multirow{4}{*}{\rotatebox{90}{PG}}
 & Centralized     & 20 - 0 - 0 & 0  & 50.29 & 261.74 \\
 & Dec. Comm.      & 20 - 0 - 0 & 0  & 50.67 & 264.21 \\
 & Dec. No Comm.   & 20 - 0 - 0 & 0  & 51.02 & 255.50 \\
 & Dec. Prediction & 17 - 0 - 3 & 11 & 46.74 & 234.85 \\
\midrule
\multirow{4}{*}{\rotatebox{90}{BGLG}}
 & Centralized     & 20 - 0 - 0 & 0  & 28.42 & 156.25 \\
 & Dec. Comm.      & 20 - 0 - 0 & 0  & 29.59 & 158.51 \\
 & Dec. No Comm.   & 20 - 0 - 0 & 1  & 30.38 & 156.61 \\
 & Dec. Prediction & 16 - 1 - 3 & 41 & 29.71 & 157.73 \\
\bottomrule
\end{tabular}
\end{center}
\end{table}

\subsection{Computational Complexity}\label{sec:sim_computation}

We finally examine how the computation time scales with the number of agents. Since the maps of the previous sections differ in the density of static obstacles, which affects the cost evaluation, we use an open crossing without static obstacles so that the number of agents is the only quantity that varies. We run the centralized variant with two to five agents, five runs each, and record the time spent solving the MPPI problem at every control cycle. The reported values cover the sampling, the rollouts, the cost evaluation, the importance sampling and the computation of the resulting input sequence. All experiments use $K=2000$ samples and run on an Intel Core Ultra 7 265H processor.

Table~\ref{tab::comp_time} reports the results. The computation time grows approximately linearly with the number of agents, which is consistent with the parallelization of the sampling procedure over the number of agents and samples. With two agents the solver meets the 10 Hz control period used throughout this work. With more agents the solve time exceeds it. For instance, with five agents, the algorithm would need to be run at a frequency below 5 Hz.

\begin{table}[t]
\caption{Solve time of the centralized variant for an increasing number of agents in an open crossing, with $K=2000$. Mean, median and standard deviation are computed over the control cycles of five runs.}
\label{tab::comp_time}
\begin{center}
\begin{tabular}{l c c c c}
\toprule
\textbf{Number of agents} & \textbf{2} & \textbf{3} & \textbf{4} & \textbf{5} \\
\midrule
Mean [ms]    & 81.1 & 128.8 & 172.4 & 211.9 \\
Median [ms]  & 76.5 & 124.2 & 169.3 & 212.0 \\
Std [ms]     & 16.8 & 26.7 & 32.4 & 50.5 \\
\bottomrule
\end{tabular}
\end{center}
\end{table}

\section{Real-World Experiments} \label{sec:realworld}

\begin{figure}[tb]
    \centering
    \includegraphics[width=0.35\textwidth,angle=270,origin=c]{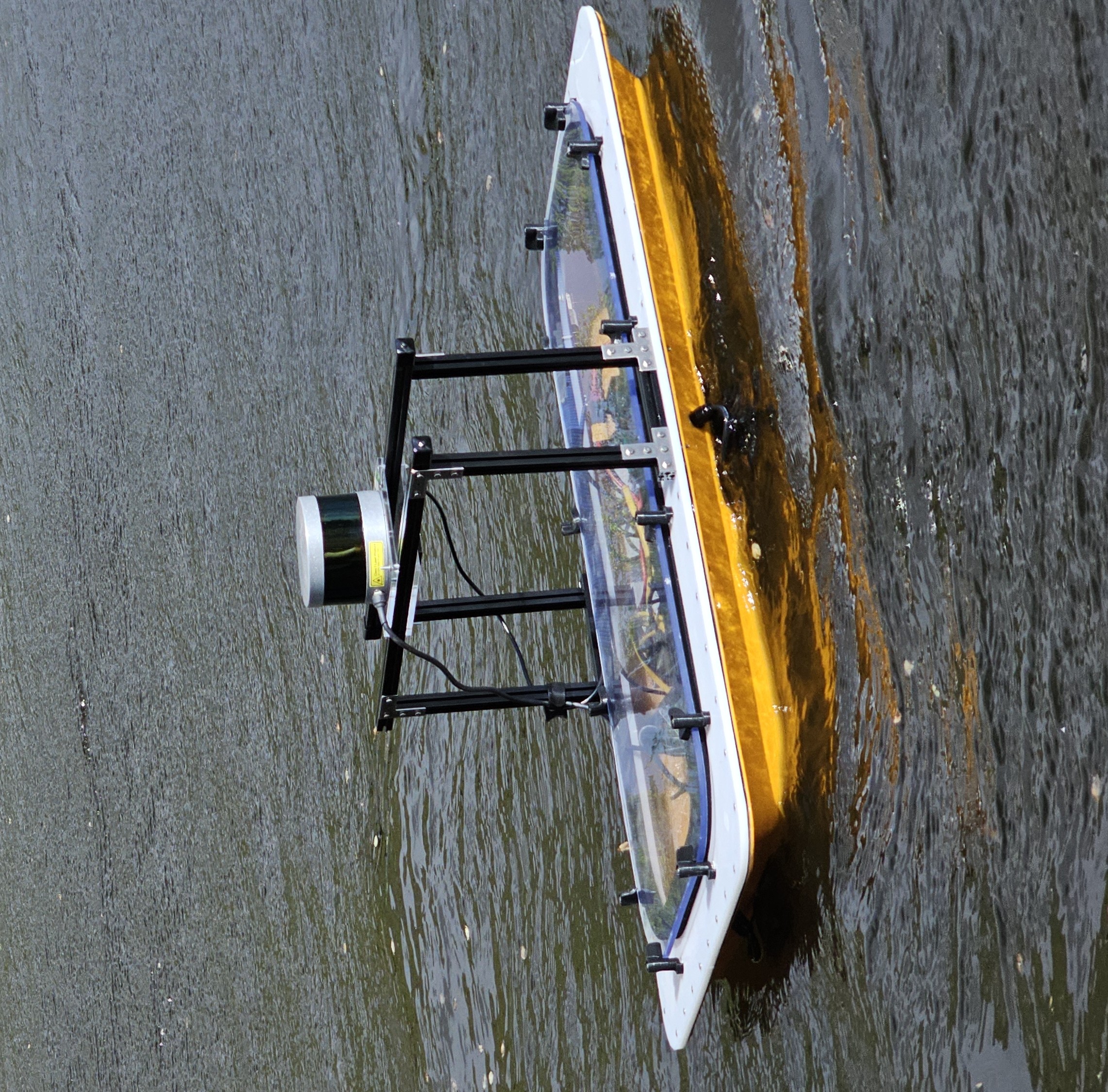}
    \caption{Quarterscale Roboat.}
    \label{img:roboat}
\end{figure}

We perform outdoor experiments using two Quarterscale Roboat platforms (Fig.~\ref{img:roboat}) \cite{wang_design_2018} at the Lake 3ME facility of TU Delft. The objective is to evaluate the proposed IA-MPPI scheme on real hardware, running onboard at a fixed control frequency. We consider the two most common maritime encounters: head-on and crossing.

\subsection{Setup}
Each Quarterscale Roboat measures 0.90 m in length, 0.45 m in width and 0.15 m in height, and weighs approximately 15 kg. The vessel is holonomic and is actuated by four Blue Robotics T200 thrusters mounted around the hull and driven by an STM32F103 microcontroller. A Microstrain 3DM-GX5 IMU and a Velodyne Puck VLP-16 3D LiDAR provide the sensing data for localization, which is fused with LIO-SAM \cite{TixiaoIROS2020a}.

All software runs onboard on an Intel NUC with an Intel Core i7-8650U processor, under Linux and the Robot Operating System (ROS). The two vessels exchange their states over WiFi, which stands in for an onboard detection and tracking system and is outside the scope of this work. No other information is transmitted: goals, plans, costs and model parameters are not shared, and each vessel predicts the local goal of the other as described in Section~\ref{sec:goalpred}.

The planner uses $K = 1000$ samples over a horizon of $T = 100$ time steps with step size $\delta T = 0.1$ s, giving a prediction horizon of 10 s, and runs at a control frequency of 10 Hz. The maximum speed is 0.3 m/s and the thrust is limited to 4 N per thruster. Each vessel is assigned a fixed target. The remaining parameters are listed in Table~\ref{tab::params}.

\subsection{Results}
We report ten head-on and ten crossing encounters. All encounters were resolved without collisions, and the vessels never incurred a rule violation. The planner sustained the 10 Hz control frequency on both vessels throughout the experiments.

\begin{figure}[t!]
    \centering
    \medskip
    \includegraphics[width=0.48\textwidth]{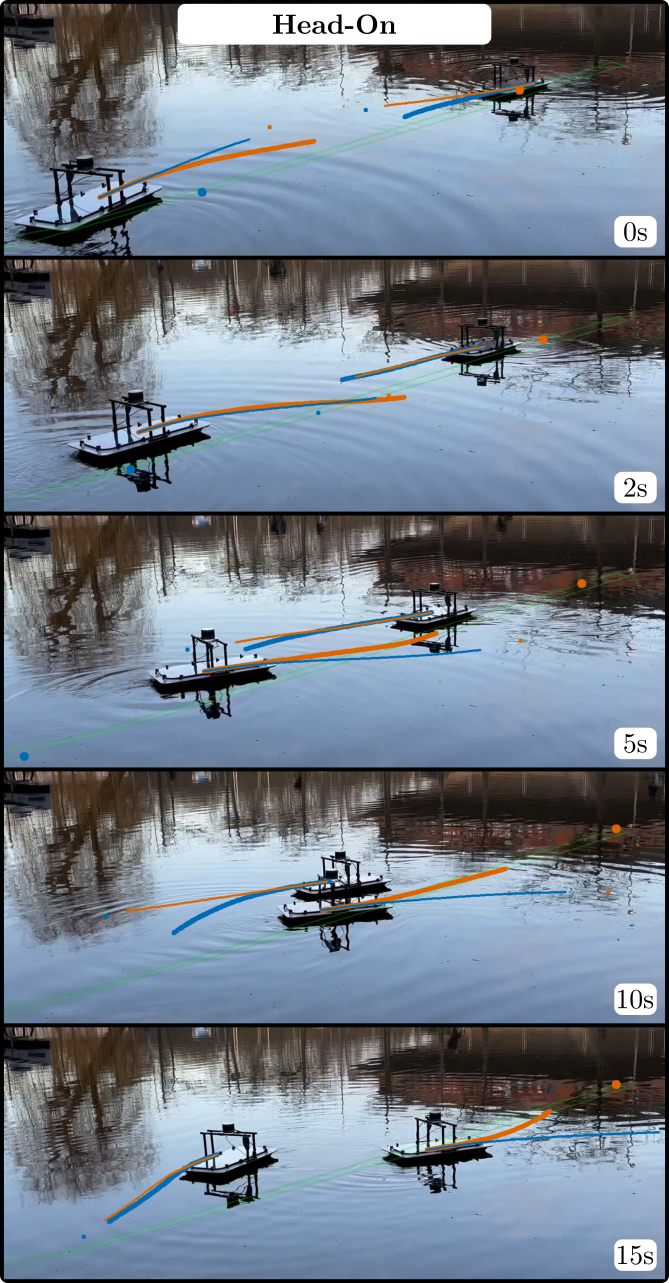}
    \caption{Two autonomous agents perform a head-on scenario using our proposed decentralized IA-MPPI. The encounter is resolved according to navigation rules. Localization and planning are performed on-board. Lines in each vessel's color are the trajectories it planned for itself and for the other vessel.}
    \label{img:headon}
\end{figure}

One of the head-on encounters is shown in Fig.~\ref{img:headon}.
At time $0s$, the vessels are still too far apart to predict an interaction within their planning horizon, and both plan to drive straight to their goals.
After two seconds, the interaction falls within the horizon and both vessels begin to alter their course to starboard, so as to pass port to port.
At five seconds, each vessel plans to perform half of the avoidance maneuver and predicts the other vessel to do the same. At time $10s$ the vessels are side by side and plan to return to their nominal course toward their goals, and at time $15s$ the encounter is resolved.

\begin{figure}[t!]
    \centering
    \medskip
    \includegraphics[width=0.48\textwidth]{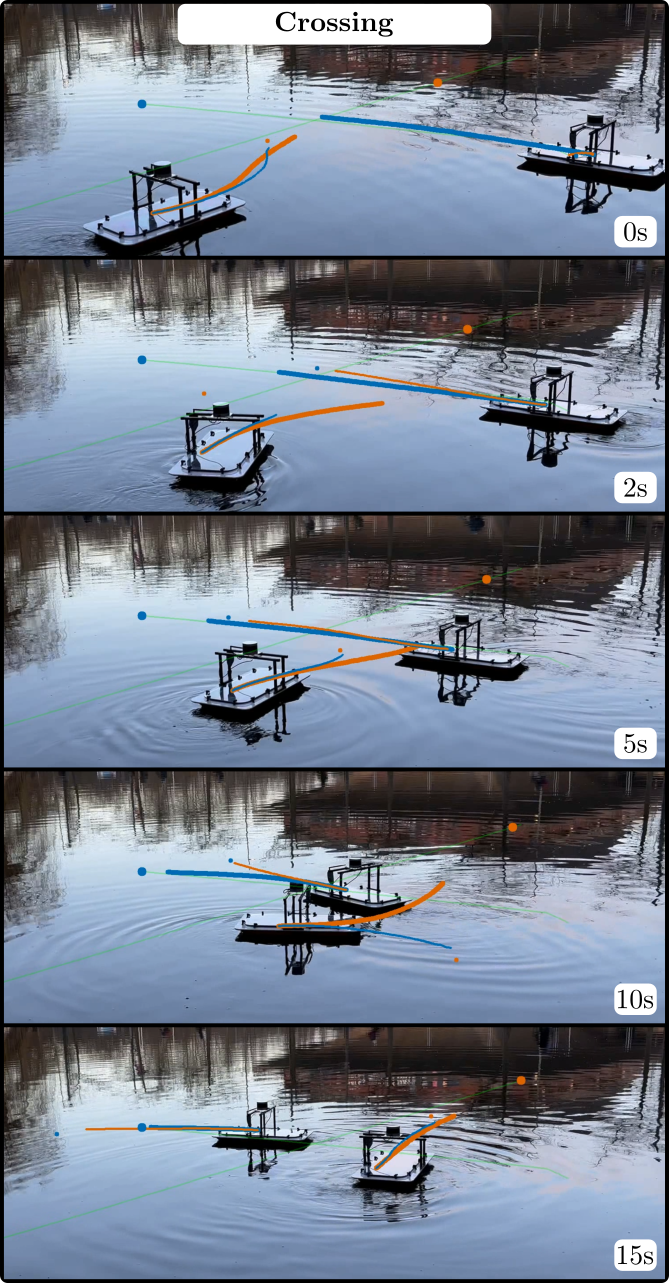}
    \caption{Two autonomous agents perform a crossing scenario using our proposed decentralized IA-MPPI. The encounter is resolved according to navigation rules. Localization and planning are performed on-board. Lines in each vessel's color are the trajectories it planned for itself and for the other vessel. }
    \label{img:crossing}
\end{figure}

One of the crossing encounters is shown in Fig.~\ref{img:crossing}.
At time $0s$, the stand-on vessel, that is, the vessel with priority, on the right side of the figure, is not yet moving. For this reason, the give-way vessel, on the left side of the figure, still plans to drive straight towards its goal.
After two seconds, the stand-on vessel has acquired some velocity, and the give-way vessel veers to starboard. At $2s$ and $5s$, the give-way vessel plans to perform most of the avoidance maneuver while predicting that the stand-on vessel keeps its course, which is how crossing encounters are to be resolved under COLREGs~\cite{johansen_ship_2016}. The stand-on vessel, conversely, plans to continue on its course and predicts that the give-way vessel will pass behind it.
At ten seconds the vessels are side by side, with the give-way vessel performing most of the avoidance maneuver, and after fifteen seconds the encounter is resolved.

\section{Discussion}\label{sec:discussion}

The simulated canal sections and the field trials point to the same conclusion for two-vessel encounters: when each vessel accounts for the interaction, exchanging states is enough, and the vessels coordinate without ever agreeing on a plan. With the decentralized variant with prediction, each vessel receives the same states but resolves the encounter on its own, and the maneuvers of the two vessels are not coordinated. With the interaction-aware variants, every simulated encounter is resolved in compliance with the rules. On the real platform, the coordination holds under the noise and drift of the onboard state estimation, the mismatch between the model in the rollouts and the real vessels, and the wind and wave disturbances present outdoors.

Coordination degrades when four vessels meet at an intersection and their goals are not communicated. Estimating the goal of a vessel with a constant velocity model assumes that it keeps its course, which does not represent the intent of a vessel that is maneuvering to resolve an encounter. A possible alternative is to predict the trajectories of the other vessels with a learned model and extract the local goals from them, which can improve the estimate in interaction-rich scenes~\cite{ia-mppi-lbm}. This would require training data representative of the traffic encountered and an inference cost compatible with onboard resources.

The formulation also assumes that all agents minimize the same cost. When this does not hold, the maneuver a vessel expects from another is not the one the other performs, either because its goal is estimated incorrectly or because it does not follow the rules. In both cases, the planner trades progress for safety. With inaccurate goal estimates, the four-vessel experiments show longer paths and occasional rule violations, but no collisions. In the case of a human-piloted vessel that disregards the rules, our previous work showed the planner coming to a stop and continuing once the vessel had passed~~\cite{streichenberg_multi-agent_2023}.

The solve time grows approximately linearly with the number of jointly planned vessels. Two vessels planning over a ten second horizon meet the 10~Hz control period both in simulation and on the embedded computer carried by the platform. However, operating in real time with larger groups would require a shorter horizon, fewer samples, or more parallel hardware. The encounters considered in this work are resolved with one to two thousand samples, so the horizon is what constrains the control period, rather than the sampling density.

The field trials use the same vessel model as the simulation, with only a difference in parameter tuning. The speed limits are lower on hardware for the stability of the platform, and the weights favor smooth maneuvers over aggressive ones. The formulation and the sampling scheme are unchanged, without explicitly accounting for practical challenges such as latency, noise and drift of the state estimate, model mismatch, or environmental disturbances. The trials were nevertheless resolved as in simulation, indicating that the method is a viable candidate for vessels operating in real urban waterways.

\section{Conclusions}\label{sec:conclusions}

We presented a decentralized interaction-aware motion planner for autonomous surface vessels in urban waterways. Each vessel plans for the local multi-agent system rather than for itself alone, which lets it anticipate the maneuvers of the others without exchanging plans, while navigation rules are encoded as discontinuous costs and the sampling distribution is informed by a set of ancillary controllers. In randomized simulations set in real sections of Amsterdam's canals, the method reached every goal without collisions or deadlocks with both two and four vessels interacting. It outperformed an optimization-based planner designed for the same domain, as well as a version of the proposed planner that does not account for the interaction.

The field experiments showed that the planner runs onboard two scaled vessel prototypes at 10 Hz over a ten second horizon, and resolves head-on and crossing encounters under the noise and drift of the onboard state estimation. The main limitation is the estimation of the intent of the other vessels: a constant velocity model is sufficient for two-vessel encounters and in wider canal sections, but not for intersections where several vessels maneuver at once. Learned trajectory predictors are a promising alternative, and would integrate into the framework without further changes.

\section*{Acknowledgment}
The authors used Claude (Anthropic) to assist in editing the text and in producing code to create the schematic figures (Figs.~\ref{img:overview}, \ref{img:regulations} and \ref{img:local_goals}). All content was reviewed and verified by the authors.


\bibliographystyle{IEEEtran}
\bibliography{IEEEabrv,cogro_mppi_bibtex}

\end{document}